%% file: ijcai23.tex
\documentclass{article}
\usepackage{ijcai23}
\input{math.tex}

\usepackage{times}
\usepackage{soul}
\usepackage{url}
\usepackage[hidelinks]{hyperref}
\usepackage[utf8]{inputenc}
\usepackage[small]{caption}
\usepackage{graphicx}
\usepackage{amsmath}
\usepackage{amsthm}
\usepackage{booktabs}
\usepackage{algorithm}
\usepackage{algorithmic}
\usepackage[switch]{lineno}
\usepackage{multirow}
\usepackage{stfloats}

\usepackage{dpr,fec}
\usepackage{setspace}

\newcommand{\ie}{\textit{i.e.}}

\newcommand{\et}{\textit{et al.}}

\title{MAT: Mixed-Strategy Game of Adversarial Training in Fine-tuning}

\author{
Zhehua Zhong $^1$\and
Tianyi Chen $^2$\And
Zhen Wang $^1$\footnote{Corresponding author.}
\affiliations
$^1$ School of Cyberspace, Hangzhou Dianzi University\\
$^2$ Microsoft
\emails
zhongzhehua@outlook.com,
tiachen@microsoft.com,
wangzhen@hdu.edu.cn
}

\begin{document}

\maketitle

\begin{abstract}
    Fine-tuning large-scale pre-trained language models has been demonstrated effective for various natural language processing (NLP) tasks. Previous studies have established that incorporating adversarial training during the fine-tuning stage can significantly enhance model generalization and robustness. However, from the perspective of game theory, such utilizations of adversarial training correspond to pure-strategy games, which are inherently limited in terms of the scope of their strategies, thereby still having room for improvement. In order to push the performance boundaries, we propose a novel \textbf{M}ixed-strategy \textbf{A}dversarial \textbf{T}raining algorithm (MAT). Methodologically, we derive the Nash equilibrium of a mixed-strategy game for adversarial training using Entropy Mirror Descent to establish MAT by sampling method. To verify the effectiveness of MAT, we conducted extensive benchmark experiments on large-scale pre-trained models, such as BERT and RoBERTa. MAT significantly outperforms the state-of-the-art methods on both the GLUE and ANLI benchmarks in terms of generalization and robustness.
\end{abstract}

\section{Introduction}
Recent years have seen significant advancements in the field of natural language processing (NLP) due to the development of large-scale pre-trained language models such as BERT \cite{DBLP:conf/naacl/DevlinCLT19}, GPT \cite{radford2018improving,radford2019language,DBLP:conf/nips/BrownMRSKDNSSAA20}, and T5 \cite{DBLP:journals/jmlr/RaffelSRLNMZLL20}. These models have achieved remarkable success across a wide range of NLP tasks, primarily because of their vast number of parameters and the ability to leverage context information. However, the complexity of these models has presented challenges in model development, training, and usage. As we know, fine-tuning is an increasingly popular and fundamental approach, which adapts a pre-trained model to a downstream task through re-training on a small set of task-specific data. Meanwhile, it has been observed that fine-tuned models are susceptible to overfitting the training data of downstream tasks, which may result in suboptimal generalization performance on unseen data points.

Recent studies \cite{DBLP:journals/corr/abs-2004-08994,DBLP:conf/iclr/ZhuCGSGL20,DBLP:conf/acl/JiangHCLGZ20,DBLP:conf/iclr/AghajanyanSGGZG21} have demonstrated that incorporating adversarial training during the fine-tuning stage can alleviate issues of overfitting and effectively improve model generalization in downstream applications. Adversarial training optimizes models by augmenting the training set with adversarial examples, which are typically generated by adding small and carefully selected perturbations to input data. However, from the perspective of game theory, the existing adversarial training methods can be mainly interpreted as pure-strategy games between the model and adversarial perturbations, which are limited in their strategy space. According to the Nash theorem \cite{10.2307/1969529}, the pure-strategy game is a subset of the mixed-strategy game, and the Nash equilibrium (which represents a stable state of the game) in the pure-strategy game may not exist. Therefore, it is natural to question \textsl{whether performance can be further improved by interpreting and reformulating adversarial training in light of the theory of the mixed-strategy game}.

\begin{figure}[ht]
    \begin{center}
        \includegraphics[width=0.45\textwidth]{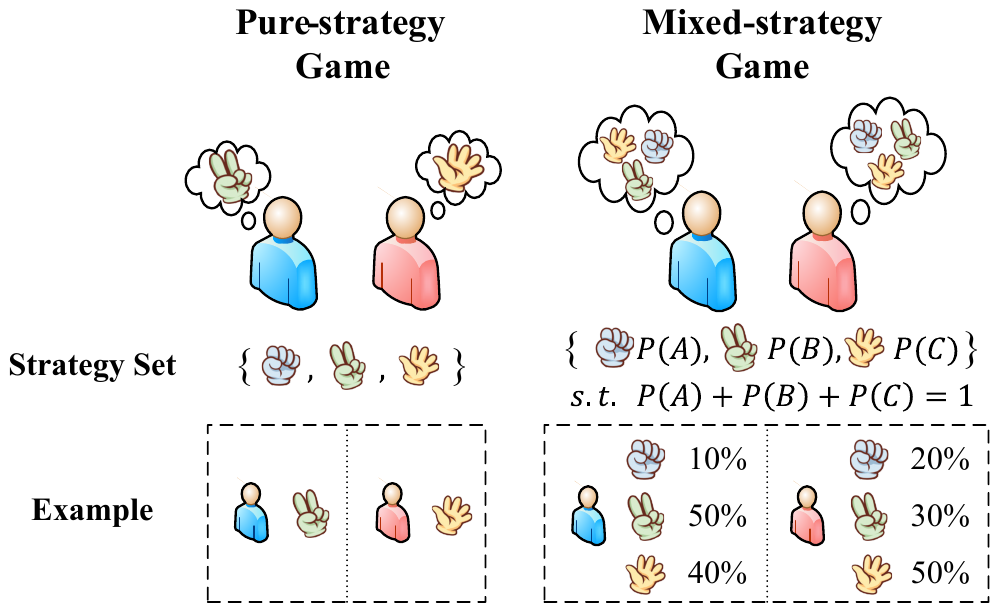}
    \end{center}
    \caption{Difference between pure strategy and mixed strategy.}
    \label{fig:figure1}
\end{figure}

In order to address the question, we view adversarial training as a two-player, complete-information game and propose an approach to improve adversarial training under the setting of a mixed-strategy game. Previous adversarial training methods can be mainly seen as pure-strategy games because both model parameters and adversarial perturbations choose deterministically specified strategies. In contrast, our proposed adversarial training method is based on the mixed strategy, which enables a broader search space for model parameters by leveraging the probabilistic nature of the strategies (as discussed in greater detail in Section \ref{Defining Adversarial Training Game}). To illustrate this concept, we can consider a two-player game of rock-paper-scissors, as shown in Figure \ref{fig:figure1}. In the mixed-strategy game, the players have a broader range of strategies to choose from.

Our main contributions can be summarized as follows:
\begin{itemize}
    \item We propose a \textbf{M}ixed-strategy \textbf{A}dversarial \textbf{T}raining algorithm (MAT) that aims to further improve the generalization and robustness of fine-tuned NLP models. Specifically, we apply game theory to interpret adversarial training as a mixed-strategy game, and derive a theoretical Nash equilibrium using the Entropy Mirror Descent method. Then we simplify this method through the use of sampling mechanisms to make the training algorithm practical and implementable in reality.
    \item We perform thorough experiments to demonstrate the effectiveness of MAT and attain state-of-the-art results on the GLUE \cite{DBLP:conf/iclr/WangSMHLB19} and ANLI \cite{DBLP:conf/acl/NieWDBWK20} benchmarks. Our findings show that applying MAT on the BERT \cite{DBLP:conf/naacl/DevlinCLT19} and RoBERTa model \cite{DBLP:journals/corr/abs-1907-11692} results in significantly superior performance compared to previous methods.
\end{itemize}

\section{Related Work}

\subsection{Adversarial Training in CV and NLP}

Adversarial training was first introduced in the field of computer vision (CV) by Szegedy \et \shortcite{DBLP:journals/corr/SzegedyZSBEGF13} and Goodfellow \et \shortcite{DBLP:journals/corr/GoodfellowSS14}. Instead of relying solely on clean datasets for training, researchers discovered that augmenting the training set with adversarial examples can effectively protect models against adversarial attacks. Madry \et \shortcite{DBLP:conf/iclr/MadryMSTV18} formulated this type of adversarial training as a Min-Max optimization problem, where the adversarial examples are generated by maximizing the deviation of the training objective within the tolerance of visibility. For solving the maximization sub-problems, the projected gradient descent (PGD) algorithm is employed to generate perturbations that satisfy the given constraints. Subsequent research by Shafahi \et \shortcite{DBLP:conf/nips/ShafahiNG0DSDTG19} and Zhang \et \shortcite{DBLP:conf/nips/ZhangZLZ019} have aimed to improve the efficiency of the PGD algorithm by reducing its time complexity.

Adversarial training in natural language processing (NLP) was introduced by Miyato \et \shortcite{DBLP:conf/iclr/MiyatoDG17}, who perturbed the word embedding space to alleviate the issue of overfitting during training models from scratch. With the advent of pre-trained large-scale models, Zhu \et \shortcite{DBLP:conf/iclr/ZhuCGSGL20} extended adversarial training to the fine-tuning stage of pre-trained models and proposed the FreeLB algorithm. FreeLB adds adversarial perturbations to word embeddings and minimizes the resultant adversarial risk inside different regions around input samples. Jiang \et \shortcite{DBLP:conf/acl/JiangHCLGZ20} developed the SMART algorithm, which effectively addresses overfitting and aggressive updating issues in the fine-tuning stage through smooth-inducing adversarial regularization and Bregman proximal point optimization. Aghajanyan \et \shortcite{DBLP:conf/iclr/AghajanyanSGGZG21} further improved upon this by decomposing pre-trained models into a feature extractor and a classifier with separate constraints and modifying adversarial perturbations to be either normal or uniformly distributed noise, significantly reducing the time required to generate adversarial perturbations.

\subsection{Game Theory Applications in Deep Learning}

Generative adversarial networks (GANs) \cite{DBLP:conf/nips/GoodfellowPMXWOCB14} generate synthetic data by mimicking genuine samples, which have become increasingly popular and widely used in various applications. GANs can be primarily interpreted as a game theory problem, which has inspired several training approaches in recent years. Arora \et \shortcite{DBLP:conf/icml/Arora0LMZ17} proposed the MIX+GAN protocol, a mixed-strategy GAN training algorithm that uses a small mixture of discriminators and generators. Daskalakis \et \shortcite{DBLP:conf/iclr/DaskalakisISZ18} introduced the Optimistic Mirror Descent (OMD) method for training Wasserstein GANs, and they formally showed that the last iteration of the OMD dynamics converged to a Nash equilibrium in the case of a zero-sum game. Hsieh \et \shortcite{DBLP:conf/icml/HsiehLC19} optimized the set of probability distributions over the pure strategies of neural networks and suggested a provably convergent GAN training approach. Ahuja \et \shortcite{DBLP:conf/icml/AhujaSVD20} formulated the standard risk minimization paradigm of machine learning in the context of finding the Nash equilibrium of an ensemble game among several environments. Although there is limited research on applying game theory to fine-tuning pre-trained models, Zuo \et \shortcite{DBLP:conf/emnlp/ZuoLJLHGCZ21} proposed the SALT approach, which formulates adversarial training as a Stackelberg game in which the follower generates perturbations and the leader optimizes model parameters given these perturbations.

\section{Mixed-Strategy Adversarial Training}

In this section, we examine adversarial training from a game-theoretic perspective. Specifically, we first generalize the mixed-strategy form of adversarial training in Section \ref{Defining Adversarial Training Game}. Then, in Section \ref{Finding Nash Equilibrium of Mixed-Strategy Adversarial Training Game}, we utilize the Entropy Mirror Descent optimization algorithm to derive a Nash equilibrium for the mixed-strategy adversarial training game. Since the probability distribution of the strategy set is implicitly unknown during model training, we propose a sampling-based simplification of the theoretical algorithm in Section \ref{From Theory to Practice}, resulting in the development of our practical mixed-strategy adversarial training algorithm, referred to as MAT.
\vspace{-0.3em}
\paragraph{Notation.} Throughout the paper, we use $f(\vtheta,\vx)$ to represent the output of the language model $f$ with parameters $\vtheta$, given input word embeddings $\vx$. $\vdelta$ represents adversarial perturbations. $\mathcal{D}_{KL}(P \| Q)=\sum_k p_k \log(p_k / q_k)$ represents the KL-divergence of two discrete distributions $P$ and $Q$.

\subsection{Defining Adversarial Training Game}\label{Defining Adversarial Training Game}

Without loss of generality, we define the loss function $L(\vtheta)$ for the standard NLP fine-tuning task as follows:
\begin{linenomath*}
\begin{equation}\label{eq:loss_fine_tune}
    L(\vtheta):=\ell\left(f\left(\vtheta,\vx\right),y\right)\text{,}
\end{equation}
\end{linenomath*}
where $y$ is the ground truth, and $\ell$ measures the deviation between the predicted output $f(\vtheta,\vx)$ and ground truth $y$. Moreover, the loss function of adversarial training is defined as:
\begin{linenomath*}
\begin{equation}\label{eq:adv_training}
    R(\vtheta,\vdelta)=\ell_s\left(f\left(\vtheta,\vx\right),f\left(\vtheta,\vx+\vdelta\right)\right)\text{,}
\end{equation}
\end{linenomath*}
where $\ell_s$ measures the deviation brought by adding adversarial perturbations onto the model inputs, which is chosen according to different tasks. For classification tasks, $\ell_s$ is typically chosen as the symmetrized KL-divergence, \ie, $\ell_s(P,Q)=\mathcal{D}_{KL}(P\Vert Q)+\mathcal{D}_{KL}(Q\Vert P)$ given two distributions $P$ and $Q$. For regression tasks, $\ell_s$ is chosen as the squared loss. Then, by combining Equation (\ref{eq:loss_fine_tune}) and Equation (\ref{eq:adv_training}), the target optimization problem for the fine-tuning tasks with adversarial training can be formulated as below:
\begin{linenomath*}
\begin{equation} \label{minmax}
    \min_{\vtheta \in \Theta} \max _{\vdelta \in \Delta}\left[L\left(\vtheta\right)+\lambda R\left(\vtheta, \vdelta\right)\right]\text{,}
\end{equation}
\end{linenomath*}
where $\lambda$ is a tuning coefficient that balances the two individual loss functions $L(\vtheta)$ and $R(\vtheta,\vdelta)$.

Equation (\ref{minmax}) can be interpreted as a potential game problem involving two parties: one for the model trainable parameters and the other one for the adversarial perturbations. The feasible domain of the model parameters can be considered as model strategy space, similar to the adversarial perturbations. Adversarial training can be solely viewed as a game where the model parameters select a strategy from the strategy set $\Theta$, while the adversarial perturbations select a strategy from the strategy set $\Delta$. Clearly, this is a pure-strategy game in which the decisions of both parties are deterministic and non-probabilistic. Existing adversarial training methods are limited to pure-strategy games with a limited strategy space, thus, we aim to extend it to a mixed-strategy game.

By using the mixed strategy, we no longer consider specific values of the model parameters and adversarial perturbations, but instead the probability of each possibility. In practice, we consider the set of all probability distributions over $\Theta$ and $\Delta$. Let $\mathcal{M}(\Theta)$ and $\mathcal{M}(\Delta)$ denote the set of all Borel probability measures on $\Theta$ and $\Delta$, respectively. Then we define the following target problem for a mixed-strategy adversarial training game to find the optimal strategy probability distribution $\mu\in \mathcal{M}(\Theta)$ and $\nu \in \mathcal{M}(\Delta)$:
\begin{linenomath*}
\begin{equation}\label{mix_minmax}
    \min _{\mu \in \mathcal{M}(\Theta)} \mathbb{E}_{\vtheta \sim \mu}\max _{\nu \in \mathcal{M}(\Delta)} \mathbb{E}_{\vdelta \sim \nu} \left[L\left(\vtheta\right)+\lambda R\left(\vtheta, \vdelta\right)\right]\text{.}
\end{equation}
\end{linenomath*}

By redefining adversarial training as a mixed-strategy game, we expand the strategy space, enabling the parties to find the optimal distribution for both sides. More strategy options allow for greater flexibility and robustness in the adversarial training process. The game can be solved when a Nash equilibrium is reached, in which no party can benefit by changing its strategy given the strategy of the other party.

\subsection{Finding Nash Equilibrium of Mixed-Strategy Adversarial Training Game} \label{Finding Nash Equilibrium of Mixed-Strategy Adversarial Training Game}

In order to determine the Nash equilibrium of the mixed-strategy adversarial training game defined in Equation (\ref{mix_minmax}), we turn to iterative optimization methods as a natural selection for solving such problems. These methods involve iteratively searching for the strategies of the players until an approximate Nash equilibrium is reached. In this work, we utilize the Entropy Mirror Descent (EMD) \cite{DBLP:journals/orl/BeckT03}, which is a specific form of the Mirror Descent (MD) \cite{bookMD}. EMD uses the entropy function as a measure of distance between probability distributions, known as the Bregman divergence. This measure is employed to guide the iterative updating of solutions, allowing us to find the Nash equilibrium of mixed-strategy games.

To proceed, we iteratively update the probability distribution of model parameters and adversarial perturbations by EMD, until an approximate Nash equilibrium is found. To simplify the notation, we abbreviate the game function defined in Equation (\ref{mix_minmax}) as $G$. The update mechanisms for the strategy sequences $\left\{\mu_t\right\}_{t=1}^T$ and $\left\{\nu_t\right\}_{t=1}^T$ can be written as: 
\begin{linenomath*}
\begin{equation}\label{eq:emd_update_mechanism}
    \left\{
    \begin{array}{l}
        \mu_{t+1}=\mathrm{EMD}_{\eta}\left(\mu_{t},\partial G / \partial \mu_{t}\right) \\
        \nu_{t+1}=\mathrm{EMD}_{\eta}\left(\nu_{t},- \partial G / \partial \nu_{t}\right)
    \end{array}  
    \quad\Rightarrow\quad
    \left(\bar{\mu}_{T}, \bar{\nu}_{T}\right)
    \text{,} 
    \right.
\end{equation}
\end{linenomath*}
where $\bar{\mu}_T$ and $\bar{\nu}_T$ are the averages of corresponding sequences, \ie, $\bar{\mu}_T:=\frac{1}{T} \sum_{t=1}^T \mu_t$ and $\bar{\nu}_T:=\frac{1}{T} \sum_{t=1}^T \nu_t$. And $ \left(\bar{\mu}_{T}, \bar{\nu}_{T}\right)$ serves as an approximate Nash equilibrium. 

Following Beck and Teboulle \shortcite{DBLP:journals/orl/BeckT03}, the EMD method replaces the Bregman divergence used in MD with an entropy function $\Phi$. This results in the following iterative updates:
\begin{linenomath*}
\begin{equation}\label{EMD}
    z_{+}=\mathrm{EMD}_{\eta}(z, h)=\nabla \Phi^{\star}(\nabla \Phi(z)-\eta h)\text{.}
\end{equation}
\end{linenomath*}

The use of the entropy function in the EMD method allows it to take into account the inherent randomness and uncertainty present in mixed-strategy games, resulting in improved convergence and stability compared to other variants of the MD method. Specifically, the Shannon entropy is used as the measure of distance when the random variable $z$ is discrete, and it is defined as $\Phi(Z) = \sum_{i=1}^n z_i \log z_i$, whereas, the differential entropy is used when $z$ is continuous and it is defined as $\Phi(Z) = \int p(z) \log p(z) \text{d}z$, where $p(z)$ is the probability density function.
To complete Equation (\ref{eq:emd_update_mechanism}), the next step is to compute the derivatives of $G$ with respect to $\mu$ and $\nu$, which are given as follows:
\begin{linenomath*}
\begin{equation} \label{partial derivatives}
    \begin{split}
        &h_{\mu}(\vtheta):=\frac{\partial G}{\partial \mu}=\mathbb{E}_{\vdelta \sim \nu}[L(\vtheta)+\lambda R(\vtheta, \vdelta)] \text{,} \\
        &h_{\nu}(\vdelta):=\frac{\partial G}{\partial \nu}=\mathbb{E}_{\vtheta \sim \mu}\lambda R(\vtheta, \vdelta) \text{.}
    \end{split}
\end{equation}
\end{linenomath*}

\begin{algorithm}[b]
\setstretch{1.2}
    \caption{EMD Adversarial Training}\label{alg_emd}
    \begin{algorithmic}[1]
        \REQUIRE Initial Distributions $\mu_1$ and $\nu_1$, Learning Rate $\eta$.
        \FOR{$t=1,2,\dots,T-1$}
        \STATE $\nu_{t+1}=\mathrm{EMD}_{\eta}\left(\nu_{t},-h_{\nu}^t\right)$
        \STATE $\mu_{t+1}=\mathrm{EMD}_{\eta}\left(\mu_{t},h_{\mu}^t\right)$
        \ENDFOR
        \STATE $\bar{\mu}_{T}=\frac{1}{T} \sum_{t=1}^{T} \mu_{t}$
        \STATE \textbf{return} $\bar{\mu}_{T}$
        \ENSURE Model Parameters Distribution $\bar{\mu}_{T}$.
    \end{algorithmic}
\end{algorithm}

To summarize the above steps, we provide an algorithmic pseudocode as Algorithm \ref{alg_emd}, which allows us to iteratively converge to the Nash equilibrium of the mixed-strategy adversarial training game theoretically. However, since the density functions of $\mu$ and $\nu$ are implicit, it is not possible to directly find the Nash equilibrium in practice. Therefore, we require a practical method to estimate these distributions, which will be discussed in Section \ref{From Theory to Practice}.

\subsection{From Theory to Practice} \label{From Theory to Practice}

Optimizing the desired risk over the distributions of model parameters and adversarial perturbation as outlined in Algorithm \ref{alg_emd} is not practical. To overcome this, the empirical risk needs to be optimized through sampling in order to obtain an approximate distribution, which requires the density function of the sampled distribution. 

Following the theorem proposed by Hsieh \et \shortcite{DBLP:conf/icml/HsiehLC19}, which illustrated an explicit form of Infinite-Dimensional Mirror Descent, we can recursively apply EMD iterations to obtain the final density functions of $\mu$ and $\nu$ as below:
\begin{linenomath*}
\begin{equation} \label{eq:density function}
\mathrm{d} \mu_{T}=\frac{e^{-\sum_{t=1}^{T} h_{\mu}^t} \mathrm{d} \vtheta}{\int e^{-\sum_{t=1}^{T} h_{\mu}^t} \mathrm{d} \vtheta}\text{,\quad}
\mathrm{d} \nu_{T}=\frac{e^{\sum_{t=1}^{T} h_{\nu}^t} \mathrm{d} \vdelta}{\int e^{\sum_{t=1}^{T} h_{\nu}^t} \mathrm{d} \vdelta}\text{.}
\end{equation}
\end{linenomath*}
Given Equation~(\ref{eq:density function}), the $\mu_t$ and $\nu_t$ can be sampled and approximated by a set of $\{\vtheta_t^{(k)}\}_{k=1}^{K}$ and $\{\vdelta_t^{(k)}\}_{k=1}^{K}$. This equation indicates that by starting sampling from $\mu_1$, it is possible to recursively estimate the distribution of all $\left\{\mu_t\right\}_{t=1}^T$.

The remaining preparation is to estimate $h_{\mu}$ and $h_{\nu}$, which are the expectations of the partial derivatives of $G$ with respect to $\mu$ and $\nu$. A commonly-used method to estimate unknown ground-truth expectations is to use the empirical average. To proceed, given $K$ samples, the partial derivatives of $\mu$ and $\nu$ in Equation (\ref{partial derivatives}) can be approximated as follows:
\begin{linenomath*}
\begin{equation} \label{partial derivatives samples}
    \begin{split}
        &\hat{h}_{\mu}(\vtheta) \approx \frac{1}{K} \sum_{k=1}^K\left[L(\vtheta)+\lambda R\left(\vtheta, \vdelta^{(k)}\right)\right] \text{,} \\
        &\hat{h}_{\nu}(\vdelta) \approx \frac{1}{K} \sum_{k=1}^K \lambda R\left(\vtheta^{(k)}, \vdelta\right) \text{.}
    \end{split}
\end{equation}
\end{linenomath*}

During model training, maintaining the entire sample set $\{\vtheta_t^{(k)}\}_{k=1}^{K}$ and $\{\vdelta_t^{(k)}\}_{k=1}^{K}$ can be computationally intensive in terms of both space complexity and time complexity. So we use the mean of samples $\bar{\vtheta}_t$ and $\bar{\vdelta}_t$ instead of all sample points. This allows us to further simplify Equation (\ref{partial derivatives samples}) as:
\begin{linenomath*}
\begin{equation}\label{eq:partial derivatives samples mean}
    \begin{split}
        &\hat{h}_{\mu}(\vtheta) \approx L(\vtheta)+\lambda R\left(\vtheta, \Bar{\vdelta}\right) \text{,} \\
        &\hat{h}_{\nu}(\vdelta) \approx \lambda R\left(\Bar{\vtheta}, \vdelta\right) \text{,}
    \end{split}
\end{equation}
\end{linenomath*}
where $\Bar{\vtheta}$ and $\Bar{\vdelta}$ are efficiently calculated by taking the exponential moving average, like $\Bar{\vz}_{t+1} \leftarrow \beta \Bar{\vz}_t+(1-\beta) \vz_t$.

Now we have only one step left, which is collecting samples from distribution $\mu$ and $\nu$. A widely used sampling method is Stochastic Gradient Langevin dynamics (SGLD) \cite{DBLP:conf/icml/WellingT11}, which consists of Stochastic Gradient Descent and Langevin Dynamics. For any probability distribution with density function $e^{-h}\mathrm{d}\vz$, such as Equation (\ref{eq:density function}), SGLD iterates as fellows:
\begin{linenomath*}
\begin{equation}\label{eq:SGLD}
\vz^{(k+1)}=\vz^{(k)}-\gamma \nabla h(\vz^{(k)})+\sqrt{2 \gamma} \epsilon \vxi \text{,}
\end{equation}
\end{linenomath*}
where $\gamma$ is the sampling step size, $ \epsilon$ is the thermal noise, and $\vxi \sim \mathcal{N}(0,1)$ is a standard normal vector.
In addition to SGLD, Li \et \shortcite{DBLP:conf/aaai/LiCCC16} demonstrated that gradient-based optimization algorithms, including RMSprop \cite{hinton2012neural} and Adam \cite{DBLP:journals/corr/KingmaB14}, can also perform dynamic sampling. Their sampling iterations are as follows:
\begin{linenomath*}
\begin{equation}
\begin{split}
    \vz^{(k+1)}&=\vz^{(k)}-\gamma \mathrm{RMSprop}\left(\nabla h(\vz^{(k)})\right)+\sqrt{2 \gamma} \epsilon \vxi \text{,}
    \\
    \vz^{(k+1)}&=\vz^{(k)}-\gamma \mathrm{Adam}\left(\nabla h(\vz^{(k)})\right)+\sqrt{2 \gamma} \epsilon \vxi \text{,} 
\end{split}
\end{equation}
\end{linenomath*}
where $\mathrm{RMSprop}$ and $\mathrm{Adam}$ refer to the specific variants of gradient calculations used in accordance with these optimization algorithms. If we substitute $\hat{h}_{\mu}$ and $\hat{h}_{\nu}$ from Equation (\ref{eq:partial derivatives samples mean}) into Equation (\ref{eq:SGLD}), the SGLD sampling iterations of $\vdelta$ and $\vtheta$ can be written as follows:
\begin{linenomath*}
\begin{equation*}
    \vdelta_{t}^{(k+1)}=\vdelta_{t}^{(k)}+\gamma_{t} \nabla_{\vdelta}\lambda R(\bar{\vtheta}_t,\vdelta_{t}^{(k)}) +\sqrt{2 \gamma_{t}} \epsilon \vxi\text{,}
\end{equation*}
\begin{equation*}
    \vtheta_{t}^{(k+1)}=\vtheta_{t}^{(k)}-\gamma_{t} \nabla_{\vtheta} \left[L(\vtheta_{t}^{(k)})+\lambda R(\vtheta_{t}^{(k)},\bar{\vdelta}_t)\right]+\sqrt{2 \gamma_{t}} \epsilon \vxi\text{.}
\end{equation*}
\end{linenomath*}

\begin{figure}[ht]
\begin{center}
\includegraphics[width=0.46\textwidth]{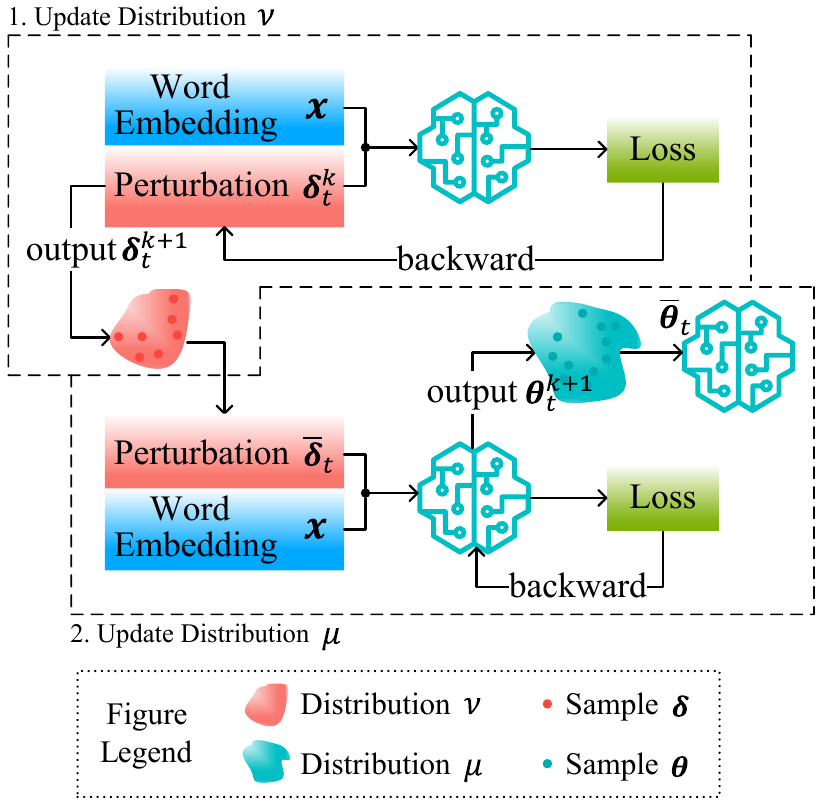}
\end{center}
\caption{Schematic diagram of MAT algorithm.}
\label{fig:figure2}
\end{figure}

In summary, we entail the substitution of the original distribution with the mean of the distribution, which was approximated through the utilization of samples drawn from said distribution. Upon procuring the approximation of the sequence $\{\mu_{t}\}_{t=1}^T$, the final Nash equilibrium is executed by calculating the average of this sequence. To provide a clear overview of our method, we outline the whole process in Algorithm \ref{alg_slgd}, which we refer to as MAT (\textbf{M}ixed-strategy \textbf{A}dversarial \textbf{T}raining). Additionally, we provide a schematic in Figure \ref{fig:figure2} to intuitively illustrate the process of MAT.

\begin{algorithm}[t]
\setstretch{1.3}
\caption{MAT: Mixed-strategy Adversarial Training}\label{alg_slgd}
\begin{algorithmic}[1]
\REQUIRE Pre-trained Model Parameters $\vtheta_1$,\\
Sampling Step Size $\{\gamma_{t}\}_{t=1}^{T}$,\\
Sampling Times $K$, Thermal Noise $\epsilon$.
\FOR{$t=1,2,\dots,T-1$}
    \STATE Initialize $\vdelta_{t}^{(1)}, \quad \bar{\vdelta}_{t}\leftarrow\vdelta_{t}^{(1)}$
    \FOR{$k=1,2,\dots,K$}
        \STATE $\vdelta_{t}^{(k+1)}\leftarrow\vdelta_{t}^{(k)}+\gamma_{t} \nabla_{\vdelta}\lambda R(\bar{\vtheta}_t,\vdelta_{t}^{(k)}) +\sqrt{2 \gamma_{t}} \epsilon \vxi$
        \STATE $\bar{\vdelta}_{t} \leftarrow \beta\bar{\vdelta}_{t}+(1-\beta)\vdelta_{t}^{(k+1)}$
    \ENDFOR
    \STATE  $\vtheta_{t}^{(1)} \leftarrow \vtheta_{t}, \quad \bar{\vtheta}_{t} \leftarrow \vtheta_{t}$
    \FOR{$k=1,2,\dots,K$}
        \STATE $\vtheta_{t}^{(k+1)}\leftarrow\vtheta_{t}^{(k)}-\gamma_{t} \nabla_{\vtheta} \left[L(\vtheta_{t}^{(k)})+\lambda R(\vtheta_{t}^{(k)},\bar{\vdelta}_t)\right]+ \sqrt{2 \gamma_{t}} \epsilon \vxi$
        \STATE $\bar{\vtheta}_{t} \leftarrow \beta\bar{\vtheta}_{t}+(1-\beta)\vtheta_{t}^{(k+1)}$
    \ENDFOR
    \STATE $\vtheta_{t+1}\leftarrow\beta\vtheta_{t}+(1-\beta)\bar{\vtheta}_{t}$
\ENDFOR
\STATE \textbf{return} $\vtheta_{T}$
\ENSURE Fine-tuned Model Parameters $\vtheta_{T}$.
\end{algorithmic}
\end{algorithm}

\section{Experiments}

In this section, in order to verify the effectiveness of our proposed algorithm, MAT\footnote{Code is available at \url{https://github.com/Charles-Zhong/MAT}.}, we conducted a comprehensive evaluation of MAT using two widely recognized natural language understanding benchmarks: GLUE \cite{DBLP:conf/iclr/WangSMHLB19} and ANLI \cite{DBLP:conf/acl/NieWDBWK20}. Additionally, MAT effectively showcase its capacity to enhance model generalization and robustness through a comprehensive comparison of experimental results with previous methods. Further details on the datasets and experiments are provided in the Appendix.

\subsection{Models and Benchmarks}

\paragraph{BERT Model.} BERT \cite{DBLP:conf/naacl/DevlinCLT19} is a transformer-based language model widely recognized for its exceptional performance on various NLP tasks. The BERT-base version, which has about 110 Million parameters, is commonly used as a base model for fine-tuning various downstream tasks.

\vspace{-0.3em}
\paragraph{RoBERTa Model.} RoBERTa \cite{DBLP:journals/corr/abs-1907-11692} is a variant of the BERT model that is trained with more extensive datasets and practical tricks such as dynamic masking, larger batch sizes, and longer training times. The RoBERTa-large version, consisting of a higher number of parameters (356M), is frequently used as a more robust base model for NLP tasks.

\vspace{-0.3em}
\paragraph{GLUE Benchmark.}  GLUE (General Language Understanding Evaluation) \cite{DBLP:conf/iclr/WangSMHLB19} benchmark is a collection of nine tasks for natural language understanding model training and evaluation. The nine datasets comprising GLUE are CoLA \cite{DBLP:journals/tacl/WarstadtSB19}, SST-2 \cite{DBLP:conf/emnlp/SocherPWCMNP13}, MRPC \cite{DBLP:conf/acl-iwp/DolanB05}, STS-B \cite{cer-etal-2017-semeval}, QQP \cite{FirstQuoraDatasetRelease}, MNLI \cite{DBLP:conf/naacl/WilliamsNB18}, QNLI \cite{DBLP:conf/emnlp/RajpurkarZLL16}, RTE \cite{DBLP:conf/mlcw/DaganGM05,bar2006second,DBLP:conf/acl/GiampiccoloMDD07,DBLP:conf/tac/BentivogliMDDG09}, and WNLI \cite{DBLP:conf/aaaiss/Levesque11}. With the exception of STS-B, which is a regression task, all of the other tasks fall within the classification category.

\vspace{-0.3em}
\paragraph{ANLI Benchmark.} ANLI (Adversarial Natural Language Inference) \cite{DBLP:conf/acl/NieWDBWK20} is an adversarial benchmark that is compiled through an iterative, adversarial human-and-model-in-the-loop process. The dataset is divided into three parts, progressively increasing the challenges for language models. ANLI is typically used to measure the robustness of models, \ie, the performance against adversarial attacks.

\begin{table*}[ht]
\begin{center}
    \begin{tabular}{c|cccccccc|c}
    \toprule
    \textbf{Dataset}  &\textbf{CoLA}  &\textbf{SST-2}  &\textbf{MRPC}  &\textbf{QQP}  &\textbf{STS-B}  &\textbf{MNLI}  &\textbf{QNLI}  &\textbf{RTE}  &\multirow{3}{*}{\textbf{Score}}
    \\ \cmidrule{1-9}
    \textbf{Size}  &8.6k  &67.3k  &3.7k &364k &5.7k &393k &104.7k  &2.5k\\
    \textbf{Metric}  &Mcc  &Acc  &Acc/F1 &Acc/F1 &P/SCorr  &Acc &Acc  &Acc
    \\ \midrule
    BERT$^1$ &-  &92.7  &86.7/-  &- &- &84.4/- &88.4  &- &- \\
    BERT$^2$ &54.7  &92.9  &84.1/89.0 &90.9/88.3 &89.2/88.8 &84.5/84.4 &91.1  &63.5 &81.48 \\
    +SMART$^2$ &59.1   &93.0  &87.7/91.3 &91.5/88.5 &90.0/89.4 &85.6/86.0 &91.7  &71.2 &83.75 \\
    +SALT$^3$ &61.0  &93.6  &88.4/91.8 &91.7/88.6 &90.4/90.0 &86.1/85.8 &92.0  &72.9 &84.48 \\
    \midrule
    +MAT (Ours)  &\textbf{62.6}  &\textbf{93.8}  &\textbf{89.0/92.1} &\textbf{91.8/89.0} &\textbf{90.5/90.2} &\textbf{86.2/86.0} &\textbf{92.4}  &\textbf{74.7} &\textbf{85.11} \\
    \bottomrule
    \end{tabular}
    \\
    \vspace{0.3em}
    References: $^1$ \cite{DBLP:conf/naacl/DevlinCLT19}, $^2$ \cite{DBLP:conf/acl/JiangHCLGZ20}, $^3$ \cite{DBLP:conf/emnlp/ZuoLJLHGCZ21}.
    \caption{GLUE results based on the BERT-base model.}
    \label{table1}
\end{center}
\end{table*}

\begin{table*}[ht]
\begin{center}
    \begin{tabular}{c|cccccccc|c}
    \toprule
    \textbf{Dataset}  &\textbf{CoLA}  &\textbf{SST-2}  &\textbf{MRPC}  &\textbf{QQP}  &\textbf{STS-B}  &\textbf{MNLI}  &\textbf{QNLI}  &\textbf{RTE}  &\multirow{3}{*}{\textbf{Score}}
    \\ \cmidrule{1-9}
   \textbf{Size}  &8.6k  &67.3k  &3.7k &364k &5.7k &393k &104.7k  &2.5k\\
    \textbf{Metric}  &Mcc  &Acc  &Acc/F1 &Acc/F1 &P/SCorr  &Acc &Acc  &Acc
    \\ \midrule
    RoBERTa$^1$ &68.0  &96.4  &90.9/- &92.2/- &92.4/- &90.2/- &94.7  &86.6 &88.93 \\
    +FreeLB$^2$ &71.1  &96.7  &91.4/-  &92.6/- &92.7/- &90.6/- &95.0  &88.1 &89.78 \\
    +SMART$^3$ &70.6  &96.9  &89.2/92.1 &92.4/89.8 &92.8/92.6 &91.1/91.3 &95.6  &\textbf{92.0} &90.09 \\
    +R3F$^4$ &71.2  &97.0  &91.6/- &92.4/89.8 &- &91.1/91.3 &95.3  &88.5 &- \\
    \midrule
    +MAT (Ours)  &\textbf{71.3}  &\textbf{97.1}  &\textbf{91.7/93.9} &\textbf{92.5/90.0} &\textbf{92.9/92.6} &\textbf{91.3/91.4}  &\textbf{95.7}  &90.6 &\textbf{90.35} \\
    \bottomrule
    \end{tabular}
    \\
    \vspace{0.3em}
    References: $^1$ \cite{DBLP:journals/corr/abs-1907-11692}, $^2$ \cite{DBLP:conf/iclr/ZhuCGSGL20}, $^3$ \cite{DBLP:conf/acl/JiangHCLGZ20}, $^4$ \cite{DBLP:conf/iclr/AghajanyanSGGZG21}.
    \caption{GLUE results based on the RoBERTa-large model.}
    \label{table2}
\end{center}
\end{table*}

\subsection{Implementation Details}

Our implementation of the MAT algorithm is based on the PyTorch framework \cite{DBLP:conf/nips/PaszkeGMLBCKLGA19}. In addition, we leverage the Huggingface library, namely \textsl{transformers} \cite{DBLP:conf/emnlp/WolfDSCDMCRLFDS20} and \textsl{datasets} \cite{DBLP:conf/emnlp/LhoestMJTPPCDPT21}, for loading pre-trained models and datasets. The text data in all the datasets are tokenized, and we limit the maximum length to 512 tokens. The hyperparameter settings are determined by combining empirical selection with the use of AutoML toolkits like NNI\footnote{\url{https://github.com/microsoft/nni}}. The checkpoints of pre-trained models, including BERT-base\footnote{\url{https://huggingface.co/bert-base-uncased}} and RoBERTa-large\footnote{\url{https://huggingface.co/roberta-large}}, are obtained from the Huggingface repository. 

During the model training phase, we utilize a range of sampling techniques, including Stochastic Gradient Langevin Dynamics (SGLD) sampling, along with its preconditioned variants such as the RMSProp-preconditioned and Adam-preconditioned versions of SGLD. Additionally, we experiment with a range of learning rates, specifically within the interval $\{1\times10^{-5}, 3\times10^{-5}, 5\times10^{-5}\}$, and employ various batch sizes, including 8, 16, 32, and 64. To ensure the robustness, we set a maximum norm of adversarial perturbation at $1\times10^{-5}$, and implement a clipping mechanism to curtail any perturbations that exceed this threshold. Furthermore, in accordance with previous works, we employ the single-task fine-tuning procedure for the GLUE benchmark, without utilizing any out-of-domain data. With regard to the ANLI benchmark, we follow the experimental setup of Nie \et \shortcite{DBLP:conf/acl/NieWDBWK20} and Jiang \et \shortcite{DBLP:conf/acl/JiangHCLGZ20} for facilitating comparison with their results, and train RoBERTa-lager model on the combined NLI datasets comprising of MNLI \cite{DBLP:conf/naacl/WilliamsNB18}, SNLI \cite{DBLP:conf/emnlp/BowmanAPM15}, FEVER \cite{DBLP:conf/naacl/ThorneVCM18}, and ANLI \cite{DBLP:conf/acl/NieWDBWK20}.

\subsection{Generalization Results}

In order to evaluate the performance of MAT, we conducted experiments on both BERT-base and RoBERTa-large models, utilizing the GLUE benchmark as the evaluation platform. The obtained results are compared to state-of-the-art adversarial training fine-tuning methods, including FreeLB \cite{DBLP:conf/iclr/ZhuCGSGL20}, SMART \cite{DBLP:conf/acl/JiangHCLGZ20}, SALT \cite{DBLP:conf/emnlp/ZuoLJLHGCZ21}, and R3F \cite{DBLP:conf/iclr/AghajanyanSGGZG21}.
Additionally, we also include several vanilla fine-tuned baseline models from Devlin \et \shortcite{DBLP:conf/naacl/DevlinCLT19}, Jiang \et \shortcite{DBLP:conf/acl/JiangHCLGZ20}, and Liu \et \shortcite{DBLP:journals/corr/abs-1907-11692} as a point of comparison. The numerical results are present in Table \ref{table1} and Table \ref{table2}, respectively, for BERT-base and RoBERTa-large models. The metrics shown in the tables are Mcc (Matthews correlation coefficient), Acc (Accuracy), F1 (F1 score), and P/SCorr (Pearson and Spearman correlation coefficient), which are 100x scaled up from percentage representations. To facilitate the comparison, the best results are highlighted in bold. Additionally, we also calculate and present the overall GLUE scores for each method, where a higher number indicates superior overall performance.  
\vspace{-0.3em}
\paragraph{BERT-base on GLUE.} As demonstrated in Table~\ref{table1}, our proposed MAT exhibits a significant improvement in model generalization compared to both vanilla fine-tuning methods and existing state-of-the-art methods such as SMART and SALT, across all individual datasets. Specifically, MAT significantly enhances the model performance on the CoLA and RTE datasets, achieving scores of 62.6 and 74.7, respectively, which represent an improvement of 1.6-7.9 and 1.8-11.2 points compared to competing methods. Furthermore, on the other datasets, MAT also demonstrates a notable superiority in performance. For instance, on large-scale datasets such as QQP and MNLI, MAT achieves the highest scores of 91.8/89.0 and 86.2/86.0, respectively, representing an improvement of 1.9/0.7 and 1.7/1.6 points over vanilla fine-tuning methods, as well as an improvement of 0.1/0.4 and 0.1/0.2 over the previous state-of-the-art method (SALT).
Overall, MAT achieves a final score of 85.11, which is substantially ahead of competitors by 0.63 to 3.63 points.
\vspace{-0.3em}
\paragraph{RoBERTa-large on GLUE.} As drawn in Table~\ref{table2}, the superiority of MAT still holds on RoBERTa-large model. MAT achieves the best performance on seven-eight datasets, with the remaining dataset ranking second. Remarkably, MAT achieves substantial improvements of 0.26-1.42 points on average and obtains the highest overall score among all methods. These results further demonstrate the effectiveness of MAT in robustly optimizing large-scale transformer models.

\subsection{Robustness Results}

Besides generalization, the robustness of models has also gained increasing attention \cite{DBLP:conf/ijcai/0002LSBLS18,DBLP:conf/aaai/JinJZS20,DBLP:conf/aaai/ZhouCZW21,DBLP:journals/compsec/WangZZYC22}. It refers to the ability of a model to maintain its performance when faced with unexpected or malicious inputs. Thus, we evaluate the robustness gained from MAT on the ANLI benchmark.

The experiments were conducted on the dev-set and test-set of ANLI, which results are in Table \ref{table3}. It is evident that MAT performs significantly better than vanilla fine-tuning \cite{DBLP:conf/acl/NieWDBWK20} and SMART \cite{DBLP:conf/acl/JiangHCLGZ20} across all levels of adversarial challenges. Specifically, MAT reaches 49.3 and 74.7 on dev-sets R3 and test-sets R1, substantially ahead of others by 1.7 and 2.3 points. These results indicate that MAT fine-tuned models exhibit superior robustness against adversarial texts across a wide range of testing scenarios.

\begin{table}[H]
\begin{center}
    \begin{tabular}{c|ccc|ccc}
    \toprule
    \multicolumn{1}{c|}{\textbf{Datasets}}&
    \multicolumn{6}{c}{\textbf{MNLI + SNLI + FEVER + ANLI}}
    \\ \midrule
    \multicolumn{1}{c|}{\textbf{Evaluation}}&
    \multicolumn{3}{c|}{\textbf{ANLI-Dev}}&
    \multicolumn{3}{c}{\textbf{ANLI-Test}}
    \\
    \textbf{ANLI Part}&  \textbf{R1}&  \textbf{R2}& \textbf{R3}&  \textbf{R1}&  \textbf{R2}&  \textbf{R3}
    \\ \midrule
    RoBERTa$^1$  &73.8  &48.9  &44.4  &-  &-  &-  \\
    +SMART$^2$  &74.5  &50.9  &47.6   &72.4  &49.8  &50.3  \\ \midrule
    +MAT (Ours)  &\textbf{74.8}  &\textbf{51.0}  &\textbf{49.3}  &\textbf{74.7}  &\textbf{51.1}  &\textbf{50.5}  \\
    \bottomrule
    \end{tabular}
    \\
    \vspace{0.3em}
    References: $^1$ \cite{DBLP:conf/acl/NieWDBWK20}, $^2$ \cite{DBLP:conf/acl/JiangHCLGZ20}.
    \caption{ANLI results based on the RoBERTa-large model.}
    \label{table3}
\end{center}
\end{table}

\begin{figure*}[t]
\begin{center}
\includegraphics[width=0.88\textwidth]{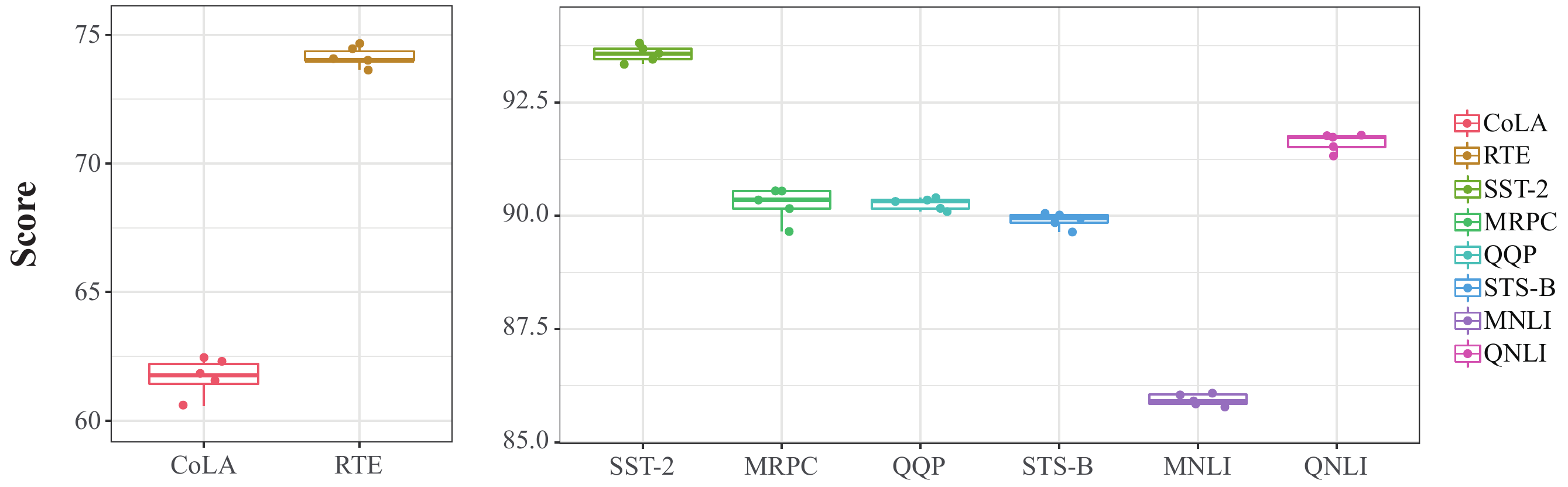}
\end{center}
\caption{Box plot error bar of MAT on the GLUE benchmark.}
\label{fig:figure3}
\end{figure*}

\subsection{Error Bar Analysis}

Training deep learning models is essentially solving a non-convex optimization problem that may have numerous local optima. Therefore, it is essential to evaluate the reliability of MAT methods by conducting sufficiently repeated experiments with different random starting points. 

To ensure fair comparisons with other methods, we conducted five experiments applying MAT to the BERT-base model on GLUE benchmarks, each with different random seeds. To visualize our experimental results, we represented them as a box plot in Figure \ref{fig:figure3}. In order to present the results more effectively, we divided all datasets into two parts, displayed on the left and right sides of the graph, with distinct vertical coordinate ranges. For datasets with multiple metrics, such as MRPC, QQP, STS-B, and MNLI, we calculated the average of these metrics to obtain the final score. The results consistently demonstrate that across all five random seeds, both the maximum and median values of our runs are consistently higher than those of other methods. These findings provide strong evidence of the reliability of MAT.

\subsection{Ablation Study}

In this subsection, we analyze the impact of several vital hyperparameters on the performance of the MAT algorithm. Specifically, we focus on the number of samples $K$ in Equation (\ref{partial derivatives samples}), which is used to approximate the probability distributions, as well as the regularization coefficient $\lambda$ in Equation (\ref{mix_minmax}) that balances the loss of regular training and adversarial training. To investigate the effect of these hyperparameters, we conducted experiments on the BERT-base model using the CoLA, SST-2, and MRPC datasets from the GLUE benchmark, while keeping all other hyperparameters constant.

The impact of the number of samples $K$ on MAT performance is shown in Table \ref{table4}. Since the exponential moving averages are leveraged in MAT, the model performance may be negatively impacted by both too small and too large values of $K$. When $K$ is too small, the distributions may not be accurately approximated. On the other hand, if $K$ is set to a large value, the estimated distributions may be overly dependent on the neighbors of the latest iterate, while ignoring other possibilities that may have occurred in previous iterates. Based on these observations, it is recommended to select an appropriate number of samples based on the coefficient of the exponential moving average to achieve the best results.

Table \ref{table5} presents the experimental results for different values of $\lambda$s. Based on our hyperparameter tuning experience, we find that choosing a larger $\lambda$ value is generally beneficial for smaller datasets. Conversely, larger $\lambda$ values tend to degrade performance on larger datasets. From the results, it can be inferred that the use of adversarial training leads to better generalization of the model over unseen samples.

\begin{table}[t]
\centering
\begin{tabular}{c|cccccc}
\toprule
\textbf{MAT-$K$} & \textbf{1} & \textbf{5} & \textbf{10} & \textbf{20} & \textbf{30} & \textbf{50} \\ \midrule
CoLA (Mcc)  &61.5	&61.7	&\textbf{62.6}	&61.6	&62.2	&61.4  \\
SST-2 (Acc) &93.1 	&\textbf{93.8} 	&\textbf{93.8} 	&93.2 	&93.0 	&93.0  \\
MRPC (Acc) &85.7	&87.0	&\textbf{89.0}	&87.3	&86.5	&86.3 \\
MRPC (F1)	&90.0	&90.9	&\textbf{92.1}	&91.1	&90.3	&90.2 \\ \bottomrule
\end{tabular}
\caption{Ablation study result of $K$ on the BERT-base model.}\label{table4}
\end{table}

\begin{table}[t]
\centering
\begin{tabular}{c|cccccc}
\toprule
\textbf{MAT-$\lambda$} & \textbf{0.1} & \textbf{1} & \textbf{3} & \textbf{5} & \textbf{10} & \textbf{50} \\ \midrule
CoLA (Mcc)  &60.8	&60.1	&\textbf{62.6}	&61.9	&62.4	&57.8  \\
SST-2 (Acc)  &93.7	&93.7	&\textbf{93.8}	&93.5	&93.5	&92.4  \\
MRPC (Acc) &86.5	&88.2	&88.7	&\textbf{89.0}	&86.3	&85.5 \\
MRPC (F1)	&90.3	&91.5	&92.0	&\textbf{92.1}  &90.2	&90.0  \\\bottomrule
\end{tabular}
\caption{Ablation study result of $\lambda$ on the BERT-base model.}\label{table5}
\end{table}

\section{Conclusion}

This work introduces a mixed-strategy game into adversarial training for fine-tuning large-scale pre-trained models, by building a game between the model and adversarial perturbations, incorporating the Entropic Mirror Descent for Adversarial Training to derive the Nash equilibrium. To simplify this method for practical usage, we propose the MAT algorithm by a sampling method. The MAT algorithm is evaluated on multiple benchmarks with pre-trained models and compared with previous state-of-the-art methods. The experimental results demonstrate that MAT leads to better performance in both model generalization and robustness. Our work also provides solid practical support for the research of introducing game theory into adversarial training.

\clearpage
\appendix
\section{Datasets}

In this supplementary section, we collate and organize the information pertaining to the datasets employed in our experiments. Table \ref{table:datasets} presents an overview of the task type, the number of classification categories,  and the sample size for each dataset. Additional details concerning the datasets are provided in the following paragraphs.

\begin{table*}[bp]
\setstretch{1.2}
\centering
\begin{tabular}{c|ccccccccc}
\toprule
\textbf{Dataset} & \textbf{CoLA}  & \textbf{SST-2} & \textbf{MRPC}  & \textbf{QQP}   & \textbf{STS-B} & \textbf{MNLI}  & \textbf{QNLI}  & \textbf{RTE}   & \textbf{ANLI}  \\ \midrule
\textbf{Task Type} & \textit{class.} & \textit{class.} & \textit{class.} & \textit{class.} & \textit{regress.}     & \textit{class.} & \textit{class.} & \textit{class.} & \textit{class.} \\
\textbf{Lables} & 2     & 2      & 2     & 2       & -     & 3       & 2       & 2     & 3       \\ \midrule
\textbf{Train-set} & 8,551 & 67,349 & 3,668 & 363,846 & 5,749 & 392,702 & 104,743 & 2,490 & 162,901 \\
\textbf{Dev-set}   & 1,043 & 872    & 408   & 40,430  & 1,500 & 19,647  & 5,463   & 277   & 3,200   \\
\textbf{Test-set}  & 1,063 & 1,821  & 1,752 & 390,965 & 1,379 & 19,643  & 5,463   & 3,000 & 3,200   \\ \bottomrule
\end{tabular}
\vspace{0.3em}
\\
Abbreviations: \textit{class.} denotes a classification task and \textit{regress.} denotes a regression task.
\caption{Summary of the datasets in the GLUE and ANLI benchmark.}
\label{table:datasets}
\end{table*}

\vspace{-0.3em}
\paragraph{CoLA Dataset.} The Corpus of Linguistic Acceptability (CoLA) is an assemblage of English acceptability judgments extracted from a corpus of literature on linguistic theory. Each example within the dataset is comprised of a sentence and accompanied by a binary label, indicating whether the sentence is deemed grammatically acceptable.

\vspace{-0.3em}
\paragraph{SST-2 Dataset.} The Stanford Sentiment Treebank (SST-2) dataset is comprised of sentences sourced from movie reviews, along with corresponding human-annotated sentiments. The task is to predict the sentiment of a given sentence, where the dataset utilizes two classification categories (positive/negative) provided labels at the sentence level.

\vspace{-0.3em}
\paragraph{MRPC Dataset.} The Microsoft Research Paraphrase Corpus (MRPC) is a corpus of sentence pairs that are automatically obtained from various online news sources and manually annotated to indicate the presence or absence of semantic equivalence within the sentence pairs.

\vspace{-0.3em}
\paragraph{QQP Dataset.} The Quora Question Pairs (QQP) dataset is a compilation of question pairs that are extracted from the community question-answering website Quora. The task of QQP dataset is to identify whether a given pair of questions are semantically equivalent or not.

\vspace{-0.3em}
\paragraph{STS-B Dataset.} The Semantic Textual Similarity Benchmark (STS-B) dataset is a collection of sentence pairs that are sourced from news headlines, video and image captions, as well as natural language inference data. Each sentence pair within the dataset is annotated by a human, which is accompanied with a similarity score from 0 to 5.

\vspace{-0.3em}
\paragraph{MNLI Dataset.} The Multi-genre Natural Language Inference corpus (MNLI) is a comprehensive dataset that is comprised of sentence pairs which are crowd-sourced and annotated with textual entailment labels. The task is to predict the relationship between a given premise sentence and a hypothesis sentence, precisely whether the premise entails the hypothesis (entailment), contradicts the hypothesis (contradiction), or neither (neutral). The premise sentences in this dataset are collected from ten different sources, including transcribed speeches, novels, and government reports.

\vspace{-0.3em}
\paragraph{QNLI Dataset.} The Question Natural Language Inference (QNLI) dataset is an adaptation of the Stanford Question Answering Dataset (SQuAD), which comprises question and paragraph pairs. In the original dataset, a sentence within a paragraph sourced from Wikipedia contains the answer to the related question, which is written by an annotator. The transformed task in QNLI is determining whether the context sentence contains the answer to the question.

\vspace{-0.3em}
\paragraph{RTE Dataset.} The Recognizing Textual Entailment (RTE) dataset comes from a series of annual textual entailment challenges, specifically RTE1, RTE2, RTE3, and RTE5. This dataset is converted into a binary classification task, where the three classes of the original datasets (neutrality, contradiction, and entailment) are collapsed into two classes (entailment and not\_entailment) for consistency.

\vspace{-0.3em}
\paragraph{ANLI Dataset.} The Adversarial Natural Language Inference (ANLI) dataset is a large-scale benchmark that is constructed through an iterative process of cyclic human-model confrontation. This dataset is divided into three parts, each of which is designed to increase the difficulty progressively for deep neural network models. ANLI is commonly utilized to evaluate the robustness of a model, specifically its ability to perform when faced with adversarial attacks.

\section{Hyperparameters}

Our experiments were conducted on multiple servers that are equipped with NVIDIA V100 and NVIDIA 3090 GPUs. As detailed in the main text of the paper, a combination of empirical methods and the AutoML toolkit (NNI) were employed for the hyperparameters search. As for the sensitivity of the hyperparameters, it is worth noting that the proposed MAT algorithm is not dependent on the meticulous tuning of the hyperparameters, as results that surpass the previous state-of-the-art were achieved with tiny hyperparameter adjustments. The value ranges of some hyperparameters that were explored during the training process are presented in Table \ref{table:Hyperparameter}.

\begin{table}[H]
\setstretch{1.1}
\centering
\begin{tabular}{c|c}
\toprule
\textbf{Hyperparameter} & \textbf{Search Range} \\ \midrule
Sampling Times $K$ & $\{5, 10, 15, 20, 30\}$  \\ 
Sampling Step Size $\gamma$ & $\{1\times10^{-5}, 3\times10^{-5}, 5\times10^{-5}\}$ \\
Beta $\beta$ & $\{0.1, 0.3, 0.5, 0.7, 0.9\}$ \\
Lambda $\lambda$ & $\{0.1, 1, 2, 3, 4,5\}$  \\\bottomrule
\end{tabular}
\caption{Search ranges of the hyperparameters.}
\label{table:Hyperparameter}
\end{table}

\clearpage

\section*{Acknowledgments}

We would like to thank Xinrun Wang and the anonymous reviewers for their helpful discussions and support. This work is supported by the Key R\&D Programs of Zhejiang (2022C01018), the Joint Funds of National Natural Science Foundation of China (U21B2001), and the National Natural Science Foundation of China (62176080).

\section*{Contribution Statements}
We state that Zhehua Zhong and Tianyi Chen contributed equally to this work.

\bibliographystyle{named}


\end{document}

%% file: math.tex
\usepackage{amsmath,amsfonts,bm}

\def\eqref#1{equation~\ref{#1}}

\def\1{\bm{1}}

\def\vtheta{{\bm{\theta}}}
\def\vdelta{{\bm{\delta}}}
\def\vxi{{\bm{\xi}}}

\def\vx{{\bm{x}}}

\def\vz{{\bm{z}}}

\DeclareMathAlphabet{\mathsfit}{\encodingdefault}{\sfdefault}{m}{sl}
\SetMathAlphabet{\mathsfit}{bold}{\encodingdefault}{\sfdefault}{bx}{n}

